# Expressing Multivariate Time Series as Graphs with Time Series Attention Transformer

**William T. Ng**[3,2*] , **K. Siu**[1] , **Albert C. Cheung**[1] and **Michael K. Ng**[2]

[1]Radiant First Research Limited
[2]The University of Hong Kong
[3]Koi Investment Partners International Limited
william.ng@koiinvestments.com, {siu.kw, albert}@radiantfirstresearch.com, mng@maths.hku.hk

## Abstract

A reliable and efficient representation of multivariate time series is crucial in various downstream machine learning tasks. In multivariate time series forecasting, each variable depends on its historical values and there are inter-dependencies among variables as well. Models have to be designed to capture both intra and inter relationships among the time series. To move towards this goal, we propose the Time Series Attention Transformer (TSAT) for multivariate time series representation learning. Using TSAT, we represent both temporal information and inter-dependencies of multivariate time series in terms of edge-enhanced *dynamic graphs*. The intra-series correlations are represented by nodes in a dynamic graph; a self-attention mechanism is modified to capture the inter-series correlations by using the super-empirical mode decomposition (SMD) module. We applied the embedded dynamic graphs to times series forecasting problems, including two real-world datasets and two benchmark datasets. Extensive experiments show that TSAT clearly outperforms six state-of-the-art baseline methods in various forecasting horizons. We further visualize the embedded dynamic graphs to illustrate the graph representation power of TSAT. We share our code at https://github.com/RadiantResearch/TSAT.

## 1 Introduction

Multivariate time series forecasting prevails in many real-world domains, such as weather forecasting, energy output management, stock prices and exchange rate predictions. In multivariate time series modeling, it is assumed that variables not only depend on their historical values but also having to take into account the latent dependencies among the variables. For example, temperatures of a city have seasonal patterns, and additionally temperatures across neighboring cities should exhibit similar seasonality.

Time series forecasting can be divided into *univariate* and *multivariate* methods. Univariate methods analyze each time series independently without cross-learning in the dataset. Traditional univariate methods are statistical methods. Variants of autoregressive integrated moving average (ARIMA), and exponential smoothing (ETS), are two prominent approaches. ARIMA models are popular because of their rich statistical foundations, and the Box-Jenkins methodology for its selection procedure [Hamilton, 1994]. For ETS, the Holt-Winters method [Winters, 1960; Holt, 2004] is one of the default models in the industry which uses an exponential smoothing technique for the level, trend, and seasonal components contained in the time series. In spite of the statistical interpretability of these methods, they lack adaptability and are limited when trying to capture non-linear relationships.

Multivariate methods consider multiple time series as a unified object; they naturally fit into the framework of deep learning methods due to their capability to use high dimensional data as inputs. Recurrent Neural Networks (RNN's) are designed for processing sequential data. For an example of RNN's application on multivariate time series see [Che *et al.*, 2016]. Another RNN technique, Long Short-Term Memory (LSTM) [Hochreiter and Schmidhuber, 1997], is designed to capture both the short and long-term patterns within sequential inputs and alleviate gradient vanishing and exploding problems in RNN models; for an example of LSTM on time series tasks see [Lai *et al.*, 2018; Shih *et al.*, 2019]. In spite of being able to capture non-linear patterns among time series, none of these methods explicitly considers the inter-series dependencies. Encoder-decoder attention models [Sutskever *et al.*, 2014; Chorowski *et al.*, 2014], and Transformer [Vaswani *et al.*, 2017], are alternatives to sequence-to-sequence learning. Especially, the self-attention mechanism that learns which entities in the input sequence are the most relevant to targeted values. Transformer has impressive results in natural language processing [Tay *et al.*, 2020], computer vision [Khan *et al.*, 2021], and ongoing efforts have been made to apply this method to time series tasks [Zhou *et al.*, 2021; Lim *et al.*, 2021]. Researchers have also attempted to design neural networks that use a mixture of basic architectures such as multi-layer perceptrons, RNNs, and Convolution Neural Networks (CNN's), for cross-learning on datasets. For example, LSTNet [Lai *et al.*, 2018] is composed of both CNN and RNN units; WaveNet [Oord *et al.*, 2016] uses dilated causal convolutions to capture long-term sequential depen-dencies. Yet, these methods cannot fully exploit the inter-relation among the time series, and hence the model interpretability is weakened.

---

[*]corresponding author

Among the deep learning methods, Graph Neural Networks (GNN's) have been successfully used to represent relationships, ranging from social network interactions, to supply chain information, to the spread of disease. A framework of graph representation learning can be summarized into three steps: 1) *Initialization step* translates the input data into a graph in terms of node features, edge features and a corresponding graph structure; 2) *Embedding step* assigns each of the graphs a unique embedding vector; 3) *Read-out step* aggregates both nodes and edge messages to represent a graph by a fixed-length output vector.

In this paper, we propose the Time Series Attention Transformer (TSAT) for multivariate time series representation learning and forecasting. First, we abstract the intra-series and inter-series correlations as a topological graph, namely *dynamic graphs*, as the temporal information is encoded and the graph structure is not static. As such, GNN can be intuitively applied to learn the representation of the dynamic graphs. Second, we attempt to integrate Transformer with GNN by augmenting the self-attention mechanism with inter-series correlation and the dynamic graph structure. The main contributions of this paper include:

- To propose the concept of an edge-enhanced dynamic graph which leads to better representation and feature engineering of multivariate time series.

- To modify the self attention mechanism for translating the time series prediction problem into graph embedding by aggregating nodes features, edges features and graph structure in one layer. Ablation studies are present to prove the effectiveness of this design.

- To provide extensive experiments on both real-world and benchmark datasets that demonstrate superior performances over the state-of-the-art methods.

The remainder of this paper is organized as follows. Section 2 lists the related work on how GNN is applied to time series forecasting. Section 3 introduces the preliminary concepts. Section 4 introduces our TSAT framework. Section 5 includes a series of experiments which we conduct and discuss. Section 6 briefly concludes the work.

## 2 Related work

**GNN for time series forecasting.** A spatial-temporal GNN assumes a physical graph structure to describe the inter-series relationships. STGCN [Yu *et al.*, 2017] assumes a traffic network and uses the graph convolutions to capture the spatial dependency among the nodes; 1D convolutions are used to model the temporal patterns. Similarly, DCRNN [Li *et al.*, 2018] uses a diffusion convolution operation and RNN units for the spatial and the temporal modelings respectively. While these methods are clearly successful, the graph structures are problem specific and require human knowledge to amend the GNNs to other problems. Another type of GNN learns the graph structures from the input time series. MT-GNN [Wu *et al.*, 2020] contains a graph learning layer which learns a graph adjacency matrix adaptively to capture the latent relations among time series data. TEGNN [Xu *et al.*, 2020] defines causality between two nodes by transfering entropy, and hence determines the connectedness. For these methods, graph structures are learnt from the state information contained as node features without fully exploiting any edge information. There are additional mathematical tools to extract features from time series before constructing the graph structure. Time2Graph [Cheng *et al.*, 2020] and [Hu *et al.*, 2021] use shapelets to capture how time series evolve over time, and hence defines evolutionary graphs. StemGNN [Cao *et al.*, 2021] projected time series into a spectral domain by using discrete Fourier transform for capturing multivariate dependencies. In our work, the graph structure is time dependent and is decided by the outputs from our SMD module.

**Transformer and GNN.** There are multiple works that integrate Transformer with GNN. Graph Transformer [Dwivedi and Bresson, 2020] modifies the original transformer to adapt to a general graph topology. The attention mechanism is converted to be a function of neighborhood connectivity for each node in the graph; the sinusoidal positional encoding is replaced by graph Laplacian eigenvectors. Graph Transformer Network (GTN) [Yun *et al.*, 2019] uses the attention mechanism to determine how to generate a meta-path between two nodes in a heterogeneous graph. A spatio-temporal GTN [Yu *et al.*, 2020] adopted graph convolution to incorporate spatial information into Transformer, allowing the model to predict a pedestrian trajectory. Molecular Attention Transformer [Maziarka *et al.*, 2020] augments the attention mechanism with inter-molecular distances and structures. Inspired by methods augmenting the self-attention mechanism with domain specific knowledge, our model encourages to improve the self-attention to adapt the node features, edge features and the graph structure of dynamic graphs.

## 3 Preliminaries

In this section, we present a novel representation learning algorithm for time series modeling. We extract features of each time series by means of super-empirical mode decomposition. The extracted feature is a set of intrinsic mode functions (IMFs) which can be expressed analytically using cosine polynomials with time-varying frequencies. We shall define a dynamic graph in terms of IMFs and incorporate the IMFs into a multi-head self-attention mechanism for learning inter-series relations.

**Super-empirical mode decomposition.** A super-empirical mode decomposition (SMD) [Chui *et al.*, 2016] is a time-analysis scheme to decompose an input time series into a finite sum of *intrinsic mode functions* (IMFs) $f_i$. i.e., given a time series $x(t)$, the SMD scheme gives

$$x(t) = \sum_{i=1}^{K} f_i(t) + R(t) \tag{1}$$

$$f_i(t) = A_i(t)\cos(2\pi\phi_i(t)). \tag{2}$$

Here, $\phi_i(t)$ refers to a non-stationary phase function and its derivative $\phi_i^{'}(t)$ is the instantaneous frequency; $A_i(t)$ is the instantaneous amplitude; $R(t)$ is a residual term which indicates a trend of the time series. In TSAT, we make use of SMD as a feature extraction module. The IMFs are key features extracted from input time series. We choose SMD over discrete Fourier transforms because the frequency contents in real-life signals are time varying.

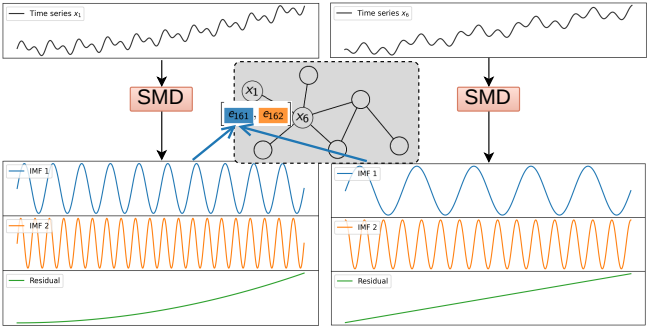

Figure 1: Each series is decomposed by SMD into a set of IMFs with a residual representing its trend. Series with similar trends are connected by an edge. Edge features are calculated by pairwise IMFs.

**Problem definition and dynamic graphs.** In multivariate time series forecasting of a rolling manner, we consider fixed lengths of backcast $L_x$ and forecast $L_y$. At time $t$, the input $\mathcal{X}^t = \{\mathbf{x}_1^t, \ldots, \mathbf{x}_{L_x}^t \mid \mathbf{x}_i^t \in \mathbb{R}^N\}$ is a set of $N$ time series of observed values and it aims to predict future values in a set $\mathcal{Y}^t = \{\mathbf{y}_1^t, \ldots, \mathbf{y}_{L_y}^t \mid \mathbf{y}_i^t \in \mathbb{R}^N\}$. Then, a dynamic graph $G^t$ is defined as a triple $(\mathbf{X}^t, \mathcal{E}^t, \mathbf{A}^t)$, where $\mathbf{X}^t \in \mathbb{R}^{N \times L_x}$ is a *node matrix* whose columns are formed sequentially from the set $\mathcal{X}^t$. Each of the nodes contains $L_x$ historical values for each time series. $\mathcal{E}^t = (e_{ijk}) \in \mathbb{R}^{N \times N \times K}$ is an *edge tensor* in which $e_{ijk}$ represents a correlation of the $k^{\text{th}}$ IMFs $f_{\cdot,k}$ between time series $x_i$ and $x_j$ given by

$$\frac{f_{i,k}^T f_{j,k}}{\|f_{i,k}\|_2 \|f_{j,k}\|_2}. \quad (3)$$

See Figure 1. Lastly, we emphasize that the graph structure is time-dependent and learnt from the input data. We use $\mathbf{A}^t \in \mathbb{R}^{N \times N}$, where $a_{ij} \in \{0, 1\}$ denotes the connectedness among the nodes. We compute the correlation of residuals of time series $x_i$ and $x_j$ by

$$\rho_{ij} = \frac{R_i^T R_j}{\|R_i\|_2 \|R_j\|_2}. \quad (4)$$

Then, the $\mathbf{A}^t$ is defined by

$$a_{ij}^t = \begin{cases} 1 & \text{if } |\rho_{ij}| > c \\ 0 & \text{otherwise.} \end{cases}$$

i.e., from SMD, the residual of a time series determines its trend. Thus, we define an edge between two nodes if their trends are highly correlated with a threshold value $c$. As such, the multivariate time series forecasting problem is formulated as a node-level supervised learning problem with additional information contained in edges.

$$\mathbf{Y}^t = F(G^t; \Theta), \quad (5)$$

where $F$ is a forecasting model with parameters $\Theta$, and $\mathbf{Y}^t$ is formed by stacking entries from the set $\mathcal{Y}^T$. As dynamic graphs cover a wide range of possible situations of multiple time series, they needs a flexible model for graph learning tasks. Our contribution is designing a modified transformer model to represent the dynamic graphs effectively.

**Attention mechanism.** The original Transformer has $M$ attention blocks. In each block there is a multi-head self-attention layer, followed by a residual connection, and a Layer Normalization [Ba *et al.*, 2016] to enhance the scalability of the model. The multi-head self-attention has $H$ heads. Denote the input matrix $\mathbf{H} \in \mathbb{R}^{n \times d}$ where $n$ and $d$ are the length and dimension of the input sequence. In each head $i (i = 1, \ldots, H)$, the operation is composed of two parts. 1) A *transformation layer* projects $X$ to three sequential matrices, namely queries, key and values respectively by

$$\mathbf{Q}_i = \mathbf{H}\mathbf{W}^{Q_i}, \quad \mathbf{K}_i = \mathbf{H}\mathbf{W}^{K_i}, \quad \mathbf{V}_i = \mathbf{H}\mathbf{W}^{V_i}, \quad (6)$$

where $\mathbf{W}^{Q_i} \in \mathbb{R}^{d \times d_k}, \mathbf{W}^{K_i} \in \mathbb{R}^{d \times d_k}$ and $\mathbf{W}^{V_i} \in \mathbb{R}^{d \times d_v}$ are three learnable weight matrices. 2) An *attention layer* computes the "attention" among the queries and keys and assigns the scores to values. The output of each of the $i$ attention operations is given by

$$\mathcal{A}_i = \sigma\left(\frac{\mathbf{Q}_i \mathbf{K}_i^T}{\sqrt{d_k}}\right) \mathbf{V}_i, \quad (7)$$

where $\sigma$ is the softmax function and the scaling factor $\sqrt{d_k}$ is present in case of gradient vanishing. All the $H$ outputs are concatenated into a single matrix $[\mathcal{A}_1, \ldots, \mathcal{A}_H] \in \mathbb{R}^{n \times H \cdot d_v}$ and then mapped by an output weight matrix $W^O \in \mathbb{R}^{H \cdot d_v \times d}$.

## 4 The framework of TSAT

In this section, we describe the architecture (Figure 2) of the Time Series Attention Transformer (TSAT). Especially, we discuss how the original self-attention mechanism is modified to capture the characteristics of time series.

**Time embedding layer.** In an ordinary Transformer, the attention mechanism neglects the order of elements in a sequence because the mechanism treats each input element simultaneously and identically. To incorporate the sequential information, a positional vector is added to each of the elements in the input sequence. Recall that the inputs of TSAT are dynamic graphs which explicitly contain temporal information. We adopt a RNN layer [Liu *et al.*, 2020] as a position embedding layer on the node matrix $\mathbf{X}^t$ which contains temporal information as

$$h(\mathbf{X}^t) = \text{RNN}(\mathbf{X}^t) \quad (8)$$

before further passing the hidden features $h(\mathbf{X}^t)$ to the next layer.

**Time series self-attention.** The original Transformer model is designed for sequence-to-sequence learning. We do not consider the dynamic graphs as sentences to be encoded. This would require additional handcrafted representations to treat dynamic graphs as strings. Instead, to aggregate information of nodes, edges and the graph structure, the multi-head self-attention operation (7) is modified as

$$\mathcal{A}_i = \left( \alpha_0 \sigma \left( \frac{\mathbf{Q}_i \mathbf{K}_i^T}{\sqrt{d_k}} \right) + \sum_{k=1}^{K} \alpha_k \sigma(\mathbf{D}_{\text{imf}_k}) + \alpha_{K+1} \mathbf{A} \right) \mathbf{V}_i, \quad (9)$$

where $\alpha_i (i = 0, \ldots, K+1)$ are trainable parameters with respect to these three pieces of information: 1) *Node features.* The first term $\mathbf{Q}_i \mathbf{K}_i^T / \sqrt{d_k}$ is an original self-attention layer which takes care of the intra-time series relation contained in nodes directly. 2) *Edge features.* There are $K$ terms for inter-series relation. $D_{\text{imf}_k}$ is a covariance matrix of the $k^{\text{th}}$ IMF of all time series. 3) *Adjacency matrix.* The last term $\mathbf{A}$ is an adjacency matrix of the dynamic graph. In a word, the TSAT operation (9) is designed for a holistic representation of a dynamic graph. It captures the node, edge, and adjacency information simultaneously in one layer.

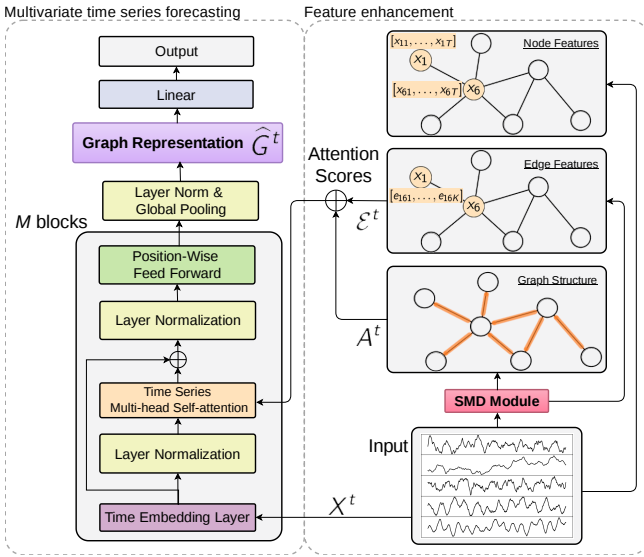

Figure 2: TSAT. The multi-head self-attention computes attention scores of three main components of a dynamic graphs, namely node features, edge features and the graph structure.

| Dataset | # samples | # nodes | sample rate |
| --- | --- | --- | --- |
| ETTh$_1$ | 17420 | 7 | hourly |
| ETTh$_2$ | 17420 | 7 | hourly |
| ETTm$_1$ | 69600 | 7 | 15-minute |
| Weather | 35064 | 1600 | hourly |
| Electricity | 26304 | 321 | hourly |

Table 1: Dataset statistics.

**Time series transformer and the read-out.** The output of TSAT layers are followed by another Layer Normalization and a position-wise Feed Forward Network (FFN) as the original Transformer encoder does. This is one block of a time series attention transformer. The TSAT block is repeated $M$ times; another Layer Normalization and a global average pooling layer are added for regularization, giving our a set of embedded dynamic graphs $\widehat{G}^t$. A fully connected layer $f(\cdot)$ is used for downstream machine learning tasks by setting $\mathbf{Y}^t = f(\widehat{G}^t)$.

## 5 Experiments

In this section, we evaluate TSAT on multivariate time series forecasting. To have a standardized comparison to other baseline methods, we adopt a similar setting to Informer [Zhou *et al.*, 2021]. We address three research questions:

- How does TSAT perform compared with state-of-the-art time series forecasting models?
- Does the formulation of dynamic graphs with nodes, edge features and the graph structure lead to better results?
- Is TSAT effective in graph representation learning for multivariate time series?

### 5.1 Experimental settings

**Benchmark datasets**
There are four datasets and five cases, including two real-world datasets and two public benchmark datasets.

**ETT (Electricity transformer temperature).** This is the same dataset as used by Informer[1]. The two-year data are recorded in two different provinces (each of seven regions) in China. The dataset is resampled as two hourly datasets, namely ETTh$_1$ and ETTh$_2$; and a fifteen-minute dataset, namely ETTm$_1$. The target value is "oil temperature".

**Weather.** This dataset was obtained from National Centers for Environmental Information[2]. It contains 1600 locations of hourly data from 2018 to 2020 in the United States. The dataset contains eleven climate features such as visibility, dew point temperature, humidity, etc. The target value is the "web-bulb" temperature.

**Electricity.** This dataset is collected from UCI machine learning repository[3] which contains hourly electricity consumption of 321 clients over two years.

All the datasets are split into 80% training data and 20% as testing data in a sequential manner. Among the training datasets, the last 10% is reserved as a validation set for the model early stopping. Detailed statistics are found in Table 1.

**Implementation details**
**Baseline models.** We selected a comprehensive set of six baseline methods composed of three categories to test against TSAT. (1) *Statistical*: ARIMA [Contreras *et al.*, 2003] is a traditional statistical model, and DeepAR [Salinas *et al.*, 2020] is a deep learning-based autoregressive model. (2) *Deep Learning*: Informer is a sparse attention based model, and LSTNet [Lai *et al.*, 2018] is composed of both CNN and RNN units. (3) *GNN based*. MTGNN [Wu *et al.*, 2020] is a graph convolutional model, and Graph WaveNet [Wu *et al.*, 2019] is a spatial-temporal graph convolution model with 1D convolution units.

---
[1] https://github.com/zhouhaoyi/ETDataset
[2] https://www.ncei.noaa.gov/data/local-climatological-data/
[3] https://archive.ics.uci.edu/ml/datasets/ElectricityLoadDiagrams20112014

**Hyper-parameter tuning.** For each method, we conduct a grid search for tunning hyper-parameters. The hyper-parameter range can be found in the Appendix. Especially, to illustrate the representation power of TSAT, the embedding dimension was chosen by $d_k = \lfloor L_x/2^n \rfloor$ for $n \in \{1, 2, 3, 4\}$. We adopt ADAM as a stochastic optimization model with a batch size equal to 64. An initial learning rate is $1e^{-4}$ and it decays exponentially with parameter $5e^{-3}$ on each epoch.

**Metrics of comparison.** For hourly datasets, the forecast horizon $L_y = \{24, 48, 168, 336, 720\}$; for the dataset of every fifteen minutes, $L_y = \{24, 48, 168, 336, 720\}$. For the input backcast length, we choose $L_x = m \times L_y (m = 1, 2, 3, 4)$. The following two evaluation metrics are used: RMSE = $\sqrt{1/n \sum_{i=1}^{N}(\mathbf{y} - \widehat{\mathbf{y}})^2}$ and MAE = $1/n \sum_{i=1}^{N} |\mathbf{y} - \widehat{\mathbf{y}}|$. To evaluate performance across the datasets, metrics are computed on the normalized samples.

## 5.2 Performance comparison

The complete results of each model on all datasets are displayed in Table 2 in which the best previous methods are underlined, and the best overall methods are shown with bold style. From the results, we observe that: (1) The proposed model TSAT has significant improvements on all datasets of different forecast horizons. TSAT records improvements in 30 out of 46 cases. For winners of each category, TSAT has better results for average RMSE than MT-GNN, LSTNet and DeepAR, by 8.02%, 16.01% and 48.28% respectively, when compared to the real world datasets $\{\text{ETTh}_1, \text{ETTh}_2, \text{ETTm}_1\}$. In the open benchmark dataset $\{\text{Weather}, \text{Electricity}\}$, the improvement in average RMSE is 6.7%, 23.66% and 50.04% over the same set of baseline methods. (2) TSAT performs better than a degraded version of TSAT, taking away both edge features and adjacency information of a dynamic graph. This indicates that both features are helpful to achieve more accurate results. (3) In two benchmark datasets $\{\text{Weather}, \text{Electricity}\}$ composed of a large number of multivariate time series (nodes), for example Electricity which has 321 time series, both TSAT and the GNN based models show superior results over the deep learning methods, and the univariate statistical methods. TSAT, MTGNN and Graph WaveNet improved by 14.66%, 13.80% and 13.47% for average RMSE on benchmark datasets, compared to real world datasets. This confirms that the graph structure can enhance the performance of tasks performed on large datasets.

## 5.3 Ablation studies

To better understand how edge features and adjacency contribute to TSAT, we include a series of ablation studies in Table 3. TSAT achieved the best results (4 out of 5 counts) when both edges and the adjacency were used as inputs. When either the edge feature or the graph structure was omitted, the performance was slightly worse than TSAT, by -1.65% and -0.57%. TSAT without both edge features and graph structure gave the worst performance. We conclude that both the use of edges features and the graph structure are essential in the performance of TSAT.

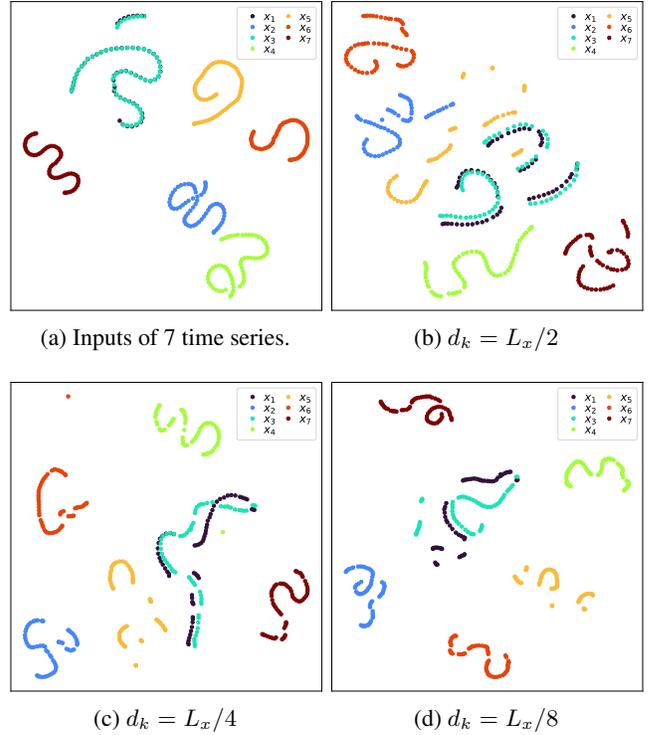

(a) Inputs of 7 time series.  (b) $d_k = L_x/2$

(c) $d_k = L_x/4$  (d) $d_k = L_x/8$

Figure 3: (a) t-SNE plots of the inputs. (b), (c) and (d) Embedded graphs of different embedding dimensions.

## 5.4 Visualization of embedded dynamic graphs

To illustrate the information preserved after using TSAT, we adopt t-distributed stochastic neighbor embedding (t-SNE) to visualize the embedded dynamic graphs of the dataset ETTh$_1$. Figure 3(a) shows a t-SNE plot composed of seven time series with length $L_x = 720$. There are clearly six clusters of points, however we observe that time series $x_1$ and $x_4$ almost overlap to each other, reflecting the similarity between these two time series in the original real world data. Figures 3(b), (c) and (d) show the t-SNE plots of TSAT outputs of different embedding dimensions ($d_k = 90, 180$ and $360$). With the additional edge features and the graph structure used by TSAT, the time series $x_1$ and $x_4$ become distinguishable. This result suggests that TSAT captures better representations of a set of multiple time series.

## 6 Conclusions

In this paper, we propose a Time Series Attention Transformer (TSAT) to translate a time series embedding problem into a graph representation problem. In the edge-enhanced dynamic graph, the inter-correlations of time series are strengthened by correlations of corresponding intrinsic mode functions which capture different cycles hidden in the input time series. The self-attention mechanism is augmented to capture a tuple of information from a graph. Extensive experiments demonstrate that our TSAT displays superior performance when compared with state-of-the-art baselines over both the real world datasets and the benchmark datasets.

| Type | Proposed model | | | | GNN based | | | | Deep learning | | | | Statistical | | | |
|---|---|---|---|---|---|---|---|---|---|---|---|---|---|---|---|---|
| Method | TSAT | | TSAT w/o graph | | MTGNN | | Graph WaveNet | | Informer | | LSTNet | | DeepAR | | ARIMA | |
| Metric | RMSE | MAE | RMSE | MAE | RMSE | MAE | RMSE | MAE | RMSE | MAE | RMSE | MAE | RMSE | MAE | RMSE | MAE |
| ETTh$_1$ 24 | **0.1864** | **0.1424** | 0.1940 | 0.1491 | 0.2002 | 0.1529 | 0.2038 | 0.1569 | 0.4828 | 0.4101 | 0.2334 | 0.1845 | 0.4940 | 0.4397 | 0.2039 | 0.1553 |
| 48 | **0.1964** | **0.1508** | 0.2065 | 0.1594 | 0.2135 | 0.1656 | 0.2189 | 0.1682 | 0.4978 | 0.4160 | 0.2205 | 0.1833 | 0.4767 | 0.4242 | 0.2101 | 0.1627 |
| 168 | 0.2283 | 0.1773 | 0.2341 | 0.1832 | 0.2505 | 0.1965 | 0.2544 | 0.1970 | 0.4861 | 0.4021 | 0.2241 | 0.1908 | 0.4662 | 0.4132 | **0.2226** | **0.1735** |
| 336 | 0.2391 | 0.1866 | 0.2402 | 0.1886 | 0.2569 | 0.2002 | 0.2632 | 0.2023 | 0.5198 | 0.4373 | 0.2626 | 0.2118 | 0.4564 | 0.4064 | **0.2307** | **0.1791** |
| 720 | 0.2492 | 0.1941 | 0.2519 | 0.1960 | 0.2687 | 0.2066 | 0.2746 | 0.2111 | 0.3176 | 0.2522 | 0.3254 | 0.2598 | 0.4911 | 0.4388 | **0.2432** | **0.1877** |
| ETTh$_2$ 24 | **0.2256** | **0.1788** | 0.2329 | 0.1837 | 0.2427 | 0.1910 | 0.2456 | 0.1941 | 0.2942 | 0.2247 | 0.2881 | 0.2205 | 0.5251 | 0.4786 | 0.3085 | 0.2206 |
| 48 | **0.2310** | **0.1810** | 0.2391 | 0.1885 | 0.2455 | 0.1938 | 0.2509 | 0.1966 | 0.3552 | 0.2757 | 0.2399 | 0.1886 | 0.4895 | 0.4420 | 0.4029 | 0.2671 |
| 168 | 0.2628 | 0.2073 | **0.2578** | **0.2032** | 0.2953 | 0.2334 | 0.3002 | 0.2361 | 0.5266 | 0.4224 | 0.5254 | 0.3321 | 0.5097 | 0.4645 | 0.4254 | 0.4144 |
| 336 | **0.2858** | **0.2251** | 0.2907 | 0.2308 | 0.3238 | 0.2534 | 0.3274 | 0.2565 | 0.5203 | 0.4197 | 0.3102 | 0.2482 | 0.5056 | 0.4599 | 3.0331 | 3.4774 |
| 720 | 0.3205 | 0.2522 | **0.3109** | **0.2471** | 0.3453 | 0.2696 | 0.3485 | 0.2745 | 0.5259 | 0.4281 | 0.5929 | 0.5288 | 0.4998 | 0.4423 | 2.4774 | 2.2194 |
| ETTm$_1$ 24 | **0.2159** | **0.1710** | 0.2177 | 0.1735 | 0.2321 | 0.1831 | 0.2344 | 0.1858 | 0.3341 | 0.2914 | 0.2714 | 0.2220 | 0.5661 | 0.5214 | 5.4671 | 4.0550 |
| 48 | **0.2386** | **0.1839** | 0.2396 | 0.1843 | 0.2554 | 0.1977 | 0.2585 | 0.1995 | 0.4561 | 0.4025 | 0.2820 | 0.2283 | 0.5310 | 0.4889 | 5.1506 | 3.8408 |
| 168 | 0.3015 | 0.2262 | **0.2983** | **0.2242** | 0.3228 | 0.2433 | 0.3265 | 0.2461 | 0.5764 | 0.4990 | 0.3049 | 0.2443 | 0.5158 | 0.4665 | 4.6019 | 3.4607 |
| 336 | 0.3467 | 0.2600 | 0.3476 | 0.2607 | 0.3714 | 0.2782 | 0.3757 | 0.2842 | 0.5766 | 0.4939 | **0.3213** | **0.2562** | 0.4962 | 0.4469 | 4.4510 | 3.3483 |
| 720 | 0.3703 | 0.2782 | 0.3658 | 0.2730 | 0.4141 | 0.3132 | 0.4276 | 0.3217 | 0.6845 | 0.6068 | **0.3393** | **0.2731** | 0.5133 | 0.4698 | 4.4076 | 3.2886 |
| Weather 48 | **0.2045** | **0.1566** | 0.2177 | 0.1687 | 0.2313 | 0.1802 | 0.2428 | 0.1854 | 0.5826 | 0.4947 | 0.4181 | 0.4649 | 0.5400 | 0.4853 | 0.2234 | 0.1596 |
| 168 | **0.2473** | **0.1898** | 0.2593 | 0.2001 | 0.2716 | 0.2099 | 0.2806 | 0.2157 | 0.6756 | 0.5713 | 0.4599 | 0.4982 | 0.5415 | 0.4817 | 0.2897 | 0.2066 |
| 336 | **0.2674** | **0.2046** | 0.2765 | 0.2113 | 0.2925 | 0.2234 | 0.2969 | 0.2266 | 0.7177 | 0.6180 | 0.4948 | 0.4250 | 0.5191 | 0.4587 | 0.9631 | 0.3099 |
| 720 | **0.2903** | **0.2222** | 0.2943 | 0.2253 | 0.3139 | 0.2396 | 0.3197 | 0.2426 | 0.7075 | 0.6044 | 0.4513 | 0.4764 | 0.5357 | 0.4726 | 0.8007 | 1.0085 |
| Electricity 12 | **0.1743** | **0.1209** | 0.1794 | 0.1248 | 0.1887 | 0.1304 | 0.1907 | 0.1322 | 0.6314 | 0.5057 | 0.3484 | 0.3585 | 0.5089 | 0.4700 | 0.4182 | 0.4593 |
| 24 | **0.1842** | **0.1258** | 0.1895 | 0.1303 | 0.1972 | 0.1344 | 0.1997 | 0.1369 | 0.5603 | 0.4550 | 0.3025 | 0.3615 | 0.5044 | 0.4653 | 0.4590 | 0.4608 |
| 48 | **0.1982** | **0.1332** | 0.2001 | 0.1352 | 0.2146 | 0.1437 | 0.2171 | 0.1479 | 0.6331 | 0.5333 | 0.3001 | 0.3590 | 0.5054 | 0.4654 | 0.5522 | 0.5693 |
| 72 | **0.2066** | **0.1361** | 0.2086 | 0.1383 | 0.2387 | 0.1562 | 0.2416 | 0.1578 | 0.6334 | 0.5981 | 0.3539 | 0.4071 | 0.5059 | 0.4668 | 0.5903 | 0.5778 |
| Count | 30 | | 6 | | 0 | | 0 | | 0 | | 4 | | 0 | | 6 | |

Table 2: Prediction results of TSTAT and 6 baselines of 3 categories on 5 cases.

| Method | ETTh$_1$ | ETTh$_2$ | ETTm$_1$ | Weather | Electricity |
|---|---|---|---|---|---|
| TSAT w/o graph | 0.2253 | 0.2663 | 0.2938 | 0.2620 | 0.1944 |
| TSAT w/o edge | 0.2208 | **0.2648** | 0.2967 | 0.2528 | 0.1910 |
| TSAT w/o adj | 0.2237 | 0.2673 | 0.2946 | 0.2595 | 0.1945 |
| TSAT | **0.2198** | 0.2651 | **0.2906** | **0.2524** | **0.1908** |

Table 3: Ablation results on 5 datasets (measured in average RMSE over different forecast horizons).

# Technical appendix

## Hyperparameter ranges

For reproducitvity of the experiments, hyperparameter ranges used in training are listed are listed for each model.

**Time series attention Transformer.** Table 4 shows hyperparameter ranges used for TSAT. Here is a list of descriptions of each parameter:

- EDGE DIM - number of IMFs used for edge features for the SMD module

- EMBED DIM - embedding dimension of the backcast horizon of time series

- ACT FUN IMF - activation function used for entries in covariance matrices of IMFs (see Equation 9)

- TSAT BLOCK - number of TSAT block repeated ($M$ used in Figure 2)

- LAYER OUTPUT - number of fully connected layers used for machine learning tasks

- ACT FUN OUTPUT - activation function that used in the LAYER OUTPUT

- DROPOUT - dropout is used between each TSAT block and the final output fully connected layer

Table 4: TSAT hyperparameter ranges.

| Item | Parameter |
| --- | --- |
| BATCH SIZE | 8, 16, 32, 64 |
| LEARNING RATE | 1e-4, decaying with 5e-3 |
| MAX EPOCH | 2000 with early stopping |
| EDGE DIM | 3, 4, 5 |
| EMBED DIM | 45, 90, 180, 256 ,360 |
| TSAT BLOCK | 1, 2, 4 ,8 |
| ATTENTION HEAD | 4, 8, 16 |
| LAYER OUTPUT | 1, 2, 4 |
| ACT FUN OUTPUT | ReLU |
| ACT FUN IMF | 'exp', 'softmax', 'none' |
| DROPOUT | 0.1, 0.2 |

**GNN based models.** MTGNN[4] and Graph WaveNet[5] are used as baseline GNN methods. Hyperparameter ranges for these methods are listed in Tables 5 and 6. For both methods, we basically followed the suggested experimental setup and provided additional parameter choices for some items to adapt to our datasets.

---

[4]https://github.com/nnzhan/MTGNN
[5]https://github.com/nnzhan/Graph-WaveNet

Table 5: MTGNN hyperparameter ranges.

| Item | Parameter |
| --- | --- |
| BATCH SIZE | 16, 32, 64 |
| LEARNING RATE | 0.001, 0.0015, 0.002 |
| EPOCHS | 100, 200, 300 |
| $l_2$ REGULARIZATION | 0.0001, 0.00015, 0.0002 |
| DROPOUT | 0.1, 0.2, 0.3 |
| GRAPH LEARNING LAYER | 2, 3 |
| MIX-HOP PRO LAYER | 2 |
| NUM SPLIT | 1,2,3 |
| NODE EMBED | 40 |

Table 6: Graph WaveNet hyperparameter ranges.

| Item | Parameter |
| --- | --- |
| BATCH SIZE | 16, 32, 64 |
| LEARNING RATE | 0.001. 0.002, 0.003 |
| WEIGHT DECAY | 0.0001, 0.0005, 0.001 |
| EPOCH | 100, 150, 200 |
| INPUT LEN | 72, 168, 720 |
| RANDOMADJ | True, False |
| NUM HID LAYER | 32, 64 |
| MODEL DROPOUT | 0.1, 0.2, 0.3 |

**Deep learning models.** Informer[6] and LSTNet[7] are used as baseline methods of deep learning. Their hyperparameter ranges are included in Tables 7 and 8.

Table 7: Informer hyperparameter ranges.

| Item | Parameter |
| --- | --- |
| BATCH SIZE | 16, 32 |
| LEARNING RATE | 0.0001, 0.005, 0.01 |
| EPOCH | 20, 30 |
| INPUT SEQUENCE LEN | 168, 720 |
| ENCODER INPUT | 7 |
| DECODER INPUT | 7 |
| MODEL DROPOUT | 0, 0.05 |
| EXPERIMENT REPEAT | 2 |

Table 8: LSTNet hyperparameter ranges.

| Item | Parameter |
| --- | --- |
| CONV1 OUT CHANNELS | 32, 64 |
| CONV1 KERNEL HEIGHT | 7 |
| RECC1 OUT CHANNELS | 32, 64 |
| SKIP STEPS | 4, 12, 24 |
| SKIP RECCS OUT CHANNELS | 2, 4, 6 |
| AR WINDOW SIZE | 7 |
| MODEL DROPOUT | 0.2 |

---

[6]https://github.com/zhouhaoyi/Informer2020
[7]https://github.com/laiguokun/LSTNet

**Statistical models.** DeepAR and ARIMA are used as baseline statistical methods. Their hyperparameter ranges are listed in Tables 9 and 10. For DeepAR, we adopted the default settings with the Electricity dataset while we included some more parameters for the grid search.

Table 9: DeepAR hyperparameter ranges.

| Item | Parameter |
| --- | --- |
| BATCH SIZE | 32, 64 |
| LEARNING RATE | 0.0001, 0.005, 0.01 |
| ENCODER | 32, 64, 168 |
| INPUT EMBED | 120, 370 |
| OUTPUT EMBED | 20, 40 |
| LSTM LAYER | 2, 3 |
| LSTM NODES | 20, 40 |

Table 10: ARIMA hyperparameter ranges.

| Item | Parameter |
| --- | --- |
| DIFFERENCING ($d$) | 0, 1 |
| MOV AVG WINDOW ($q$) | 0, 5, 10 |